\documentclass[]{article}

\usepackage{spconf,amsmath,graphicx}

\usepackage{endnotes}

\title{
Neighborhood selection for thresholding-based subspace clustering 
}
\name{Reinhard Heckel, Eirikur Agustsson, and Helmut B\"olcskei
\thanks{Part of the results in this paper were submitted 
to the Annals of Statistics \cite{heckel_robust_2013}. }
}

\address{Dept. IT \& EE, ETH Zurich, Switzerland}

\usepackage{setspace}

\usepackage{amssymb}

\usepackage[figuresright]{rotating}

\usepackage{amsmath,amssymb}
\usepackage{graphicx}
\usepackage{dcolumn}
\usepackage{bm}
\usepackage{amscd,amsthm}
\usepackage{ifthen}

\usepackage{mathtools} 

\usepackage[applemac]{inputenc}

\usepackage{supertabular}
\usepackage{booktabs}

\usepackage{tikz}

\usepackage{pgfplots}
\usepackage{subfigure}
\usetikzlibrary{pgfplots.groupplots}

\usepackage{./rays_defs}

\pgfplotsset{colormap/cool}

\usepackage{enumitem}
\usepackage{bm}

\usepackage{hyperref}
\hypersetup{
    bookmarks=true,         
    unicode=false,          
    pdftoolbar=true,        
    pdfmenubar=true,        
    pdffitwindow=false,     
    pdfstartview={FitH},    
    pdfauthor={Reinhard Heckel},     
    pdfsubject={Subject},   
    pdfcreator={Reinhard Heckel},   
    pdfproducer={Producer}, 
    pdfnewwindow=true,      
    colorlinks=true,       
    linkcolor=red,          
    citecolor=green,        
    filecolor=magenta,      
    urlcolor=cyan           
}

\definecolor{DarkBlue}{rgb}{0.1,0.1,0.5}
\definecolor{BrickRed}{RGB}{203,65,84}

\newcommand{\US}[1]{\mathbb S^{#1-1}} 

\newcommand\PR[1]{\ensuremath{ {\mathrm{P}}\!\left[#1\right]}}

\newcommand\Tex{}
\newcommand\EX[2][\Tex]{
\ifthenelse{\equal{#1}{}}{{\mathbb E}\left[#2\right]}{\ensuremath{{\mathbb E}_{#1}\left[ #2\right]}}}

\newcommand\Var[2][\Tex]{
\ifthenelse{\equal{#1}{}}{{\mathrm{Var} }[#2]}{\ensuremath{\mathrm{Var}_{#1}\left[ #2\right]}}}

\newcommand\ignore[1]{}

\newcommand\defeq{\coloneqq}

\newcommand{\reals}{\mathbb R} 
 
\newtheorem{lemma}{Lemma}

\newtheorem{theorem}{Theorem}

\newtheorem*{noiselessTSC}{Modified TSC algorithm}

\newcommand{\pinv}[1]{  {#1}^{ \dagger } } 

\newcommand{\herm}[1]{{#1}^T} 

\newcommand{\mintwo}[2]{ #1 \wedge #2 } 

\renewcommand{\d}{d} 

\newcommand{\aff}{\mathrm{aff}}

\newcommand{\cS}{S} 
\renewcommand{\d}{d} 
\newcommand{\X}{\mathcal X} 

\usepackage{bm}

\newcommand{\va}{\vect{a}}

\renewcommand{\ve}{\vect{e}}

\newcommand{\vv}{\vect{v}}  

\newcommand{\vx}{\vect{x}}  
  
\newcommand{\vz}{\vect{z}}

\newcommand{\mA}{\vect{A}}

\newcommand{\mI}{\vect{I}}

\newcommand{\mU}{\vect{U}}

\newcommand{\mX}{\vect{X}}

\newcommand{\mZ}{\vect{Z}}

\renewcommand{\S}{\mathcal T}

\newcommand{\q}{q} 
\newcommand{\s}{s}  
 
 \newcommand{\J}{\mathcal T}

\renewcommand{\l}{\ell} 

\newcommand\acos{\mathrm{arccos}}

\begin{document}
\maketitle

\begin{abstract}
Subspace clustering refers to the problem of clustering high-dimensional data points into a union of low-dimensional linear subspaces, where the number of subspaces, their dimensions and orientations are all unknown. 
In this paper, we propose a variation of the recently introduced thresholding-based subspace clustering (TSC) algorithm, which applies spectral clustering to an adjacency matrix constructed from the nearest neighbors of each data point with respect to the spherical distance measure. The new element resides in an individual and data-driven choice of the number of nearest neighbors. 
Previous performance results for TSC, as well as for other subspace clustering algorithms based on spectral clustering, come in terms of an intermediate performance measure, which does not address the clustering error directly. Our main analytical contribution is a performance analysis of the modified TSC algorithm (as well as the original TSC algorithm) in terms of the clustering error directly. 
 
\end{abstract}


\section{Introduction}

Suppose we are given a set of $N$ data points in $\reals^m$, denoted by $\X$, and assume that 
$
\X = \X_1 \cup ...  \cup  \X_L
$ 
where the points in $\X_\l, \l \in \{1,...,L\}$, satisfy $\vx_j^{(\l)} \in \cS_\l$ with $\cS_\l$ a $d_\l$-dimensional subspace of $\reals^m$. 
The association of the points in $\X$ with the $\X_\l$, the number of subspaces $L$, their dimensions $d_\l$, and their orientations are all unknown. We want to  find the partitioning of the points in $\X$ into the  sets $\X_1,...,\X_L$. Once this partitioning has been identified, it is straightforward to extract the subspaces $\cS_\l$ through principal component analysis (PCA). 
This problem is known as subspace clustering and has applications in, e.g., unsupervised learning, image processing, disease detection, and computer vision \cite{vidal_subspace_2011}. 

Numerous approaches to subspace clustering are available in the literature, see \cite{vidal_subspace_2011} for an excellent overview. 
Several recently proposed subspace clustering algorithms such as sparse subspace clustering (SSC) \cite{elhamifar_sparse_2009,elhamifar_sparse_2013}, low-rank representation (LRR) \cite{liu_robust_2010}, SSC-orthogonal matching pursuit (OMP) \cite{dyer_greedy_2013}, and thresholding-based subspace clustering (TSC) \cite{heckel_robust_2013} are based on the principle of applying spectral clustering \cite{luxburg_tutorial_2007} to a similarity matrix $\mA \in \reals^{N\times N}$ constructed from the data points in $\X$. 
Specifically, in SSC $\mA$ is obtained by finding a sparse representation of each data point in terms of the other data points via $\ell_1$-minimization (or via LASSO \cite{soltanolkotabi_robust_2013}), SSC-OMP replaces the $\ell_1$-step in SSC by OMP,  LRR computes $\mA$ through a low-rank representation of the data points obtained by nuclear norm minimization, and TSC constructs $\mA$ from  the nearest neighbors of each data point through thresholding of the correlations between data points. 

A common feature of SSC, SSC-OMP, and TSC is that $\mA$ is constructed by sparsely representing each data point in terms of all the other data points. The sparsity level of the corresponding representation is controlled by a stopping criterion for SSC-OMP, by the number of nearest neighbors for TSC, and by the LASSO regularization parameter $\lambda$ for the robust version of SSC \cite{soltanolkotabi_robust_2013}. 
A procedure for selecting $\lambda$ for each data point individually and in a data-driven fashion is described in  \cite{soltanolkotabi_robust_2013}. 


\paragraph*{Contributions:} We consider a variation of TSC---refer\-red to as ``modified TSC'' henceforth---which selects the number of nearest neighbors of each data point individually and in a data-driven fashion. 
For a semi-random data model with deterministic subspaces and the data points within the subspaces chosen randomly, we provide performance guarantees in terms of the clustering error, defined as the fraction of misclassified points. 
Specifically, we build on the fact that the clustering error is zero if the connected components\footnote{We say that a subgraph $H$ of a graph $G$ is connected if any two nodes in $H$ can be joined by a path such that all intermediate nodes also lie in $H$. The subgraph $H$ is called a connected component if $H$ is connected and if there are no connections between nodes in $H$ and nodes outside of $H$ \cite{luxburg_tutorial_2007}. 
} in the graph $G$ with adjacency matrix $\mA$ 
 correspond to the $\X_\l$. 
The performance results in \cite{soltanolkotabi_geometric_2011,soltanolkotabi_robust_2013,liu_robust_2010,dyer_greedy_2013,heckel_subspace_2013} are all based on an intermediate, albeit sensible, performance measure guaranteeing that the nodes in $G$ corresponding to $\X_\l$ are connected to other points in $\X_\l$ only, for each $\l$. This is, however, not sufficient to conclude that the connected components in the graph $G$ correspond to the $\X_\l$. 
The key to deriving conditions for TSC to yield zero clustering error is to recognize that $G$ is a random nearest neighbor graph and to analyze its connectivity properties. 
%

\vspace{-0.3cm}
\paragraph*{ Notation:} 
We use lowercase boldface letters to denote (column) vectors and uppercase boldface letters to designate matrices. For the vector $\vx$, $x_q$ stands for its $q$th entry. For the matrix $\mA$, $\mA_{ij}$ is the entry in its $i$th row and $j$th column, $\pinv{\mA}$ its pseudo-inverse, $\norm[2\to 2]{\mA} \defeq \max_{\norm[2]{\vv} = 1  } \norm[2]{\mA \vv}$ its spectral norm, and $\norm[F]{\mA} \defeq \sqrt{\sum_{i,j} |\mA_{ij}|^2 }$ its Frobenius norm.  
 $\log(\cdot)$ stands for the natural logarithm, $\acos(\cdot)$ for the inverse function of $\cos(\cdot)$, and $x \land y$ is the minimum of $x$ and $y$.  
The set $\{1,...,N\}$ is denoted by $[N]$ and the cardinality of the set $\S$ is $|\S|$. 
We write $\mathcal N( \boldsymbol{\mu},\boldsymbol{\Sigma})$ for a Gaussian random vector with mean $\boldsymbol{\mu}$ and covariance matrix $\boldsymbol{\Sigma}$. 
The unit sphere in $\reals^m$ is $\US{m} \defeq \{ \vx \in \reals^m \colon \norm[2]{\vx} = 1 \}$. 

\section{The modified TSC algorithm}

We next present a variation of the TSC algorithm introduced in \cite{heckel_subspace_2013,heckel_robust_2013}. 
The new element here is a \emph{data-driven} choice of the number of nearest neighbors  for each data point \emph{individually}. 
For Step 1 below to make sense, we assume that the data points in $\X$ are normalized. This assumption is not restrictive as the data points can be normalized prior to clustering.

\begin{noiselessTSC}
\label{alg:TSC}
Given a set of $N$ data points $\X$ and a threshold parameter $\tau$ (the choice of $\tau$ is discussed below), 
perform the following steps:

{\bf Step 1:} For every $\vx_j \in \X$, sort $\left| \innerprod{\vx_j}{ \vx_i} \right|, i \in [N]$, in descending order, and let $\S_j(q) \subseteq [N] \!\setminus\! j$ be the index set corresponding to the $q$ largest values of $\left| \innerprod{\vx_j}{ \vx_i} \right|$.  Next, determine $q_j$ as the smallest value of $q$ such that 
\begin{align}
\norm[2]{ (\mI  - \mX_{\S_j(q)}  \pinv{\mX}_{\S_j(q)} ) \vx_j } \leq \tau
\label{eq:approxadf}
\end{align}
where $\mX_{\S_j(q)}$ is the matrix with columns $\vx_i, i \in \S_j(q)$. 

{\bf Step 2:} For each $j\in [N]$, set the entries of $\vz_j \in \reals^N$ indexed by ${\S_j}(q_j)$ to the absolute values of  $\pinv{\mX}_{\S_j(q_j)}  \vx_j$ and set all other entries to zero. 
Construct the adjacency matrix $\mA$ according to $\mA = \mZ + \herm{\mZ}$, where $\mZ = [\vz_1,...,\vz_N]$. 

{\bf Step 3:} Estimate the number of subspaces as the number of zero eigenvalues, $\hat L$, of the normalized Laplacian of the graph with adjacency matrix $\mA$. 

{\bf Step 4:} Apply normalized spectral clustering \cite{luxburg_tutorial_2007} to $(\mA, \hat L)$. 
\end{noiselessTSC}

Since $\acos(z)$ is decreasing in $z$ for $z \in [0,1]$, $\S_j(q)$ is the set of $q$ nearest neighbors of $\vx_j$ with respect to the pseudo-distance metric\footnote{
$\tilde \s (\vx_i,\vx_j) =   \acos(\left|\innerprod{\vx_j}{\vx_i}\right|)$ is not a distance metric since $\tilde s(\vx,-\vx) = 0$, but $-\vx \neq \vx$ for $\vx\neq \mathbf{0}$. It satisfies, however, the defining properties of a pseudo-distance metric \cite{kelley_general_1975}.} $\acos(\left|\innerprod{\vx_j}{\vx_i}\right|)$. 
The hope is that $\S_j(q_j)$, corresponding to $\vx_j \in \X_\l$, contains points in $\X_\l$ only. 
In addition, 
we want the points corresponding to $\X_\l$, for every $\l$, to form a connected component in the graph $G$ with adjacency matrix $\mA$. If this is, indeed, the case, then by virtue of the number of zero eigenvalues of the Laplacian of $G$ being equal to the number of connected components in $G$ \cite{luxburg_tutorial_2007}, Step 3 delivers the correct estimate $\hat L = L$ for the number of subspaces. The spectral clustering Step 4 will then identify the individual connected components of $G$ and thus yield correct segmentation of the data \cite[Prop.~4; Sec.~7]{luxburg_tutorial_2007}. 
When the points corresponding to the
$\X_\l$ do not form connected components in $G$ 
but the $\mA_{ij}$ for pairs $\vx_i,\vx_j$ belonging to different $\X_\l$ are ``small enough'', a robust estimator for $L$ is the \emph{eigengap heuristic} \cite{luxburg_tutorial_2007}. With this modification, TSC may still cluster  the data correctly, even when points corresponding to, say, $\X_\l$, are connected to points in the set $\X \!\setminus \!\X_\l$. 

The idea underlying Step 1 in the modified TSC algorithm is to estimate $q_j$ as the number of points necessary to represent $\vx_j \in \X_\l$ as a linear combination of its nearest neighbors; the left-hand side of \eqref{eq:approxadf} is the corresponding $\ell_2$-approximation error. The estimate for $\q_j$ will be on the order of $d_\l$, the dimension of $\cS_\l$, the subspace $\vx_j$ lies in. To see this, assume that  the data points in $\X_\l$ are distributed uniformly at random on the set $\{\vx \in \cS_\l \colon \norm[2]{\vx} = 1 \}$. If the points corresponding to $\S_j(d_\l)$ are all in $\X_\l$, then those points suffice (with probability one) to represent $\vx_j$ with zero error. 
 Moreover, with probability one, every strict subset of these points will fail to represent $\vx_j$ with zero error. Thus, the estimate $\q_j$ obtained for $\tau = 0$ in Step 1 is equal to $\d_\l$, with probability one. 
Throughout this paper, we set $\tau=0$ in \eqref{eq:approxadf}; in the noisy case, not considered here, a sensible choice is to take $\tau$ proportional to the noise variance. 
%


\vspace{-0.1cm}
\section{Analytical performance results}
\vspace{-0.1cm}

We take the subspaces $\cS_\l$ to be deterministic and choose the points within the $\cS_\l$ randomly. To this end, we represent the  points in $\cS_\l$ by  
$
\vx_j^{(\l)} = \mU^{(\l)} \va^{(\l)}_j
$
where $\mU^{(\l)} \in \reals^{m\times \d_\l}$ is an orthonormal basis for the $\d_\l$-dimensional subspace $\cS_\l$ and the $\va^{(\l)}_j \in \reals^{\d_\l}$ are i.i.d.~uniformly distributed on $\US{\d_\l}$. Since each $\mU^{(\l)}$ is orthonormal, the data points $\vx_j^{(\l)} = \mU^{(\l)} \va^{(\l)}_j$ are uniformly distributed on the set $\{\vx \in \cS_\l \colon \norm[2]{\vx} = 1 \}$. 
Our performance guarantees are expressed in terms of the affinity between subspaces, defined as 
\begin{align}
\aff(\cS_k,\cS_\l) \defeq \frac{1}{\sqrt{ \mintwo{d_k}{d_\l} }} \norm[F]{ \herm{\mU^{(k)}} \mU^{(\l)}  }.
\label{eq:defaff}
\end{align}
Note that the affinity notion \cite[Definition~2.6]{soltanolkotabi_geometric_2011} and \cite[Definition~1.2]{soltanolkotabi_robust_2013}, relevant to the analysis of SSC, is equivalent to \eqref{eq:defaff}. 
The affinity between subspaces can be expressed in terms of the principal angles between  $\cS_k$ and $\cS_\l$ according to 
\begin{align}
\aff(\cS_k,\cS_\l) = \frac{\sqrt{ \cos^2( \theta_1) + ...+ \cos^2(\theta_{ \mintwo{d_k}{d_\l} })}}{\sqrt{\mintwo{d_k}{d_\l} }}
\label{eq:affpa}
\end{align}
where $\theta_1, ... ,\theta_{ \mintwo{d_k}{d_\l}}$ with $0 \leq \theta_1 \leq ... \leq  \theta_{ \mintwo{d_k}{d_\l} }\leq \pi/2$ 
denotes the principal angles \cite[Sec.~12.4.3]{golub_matrix_1996} between $\cS_k$ and $\cS_\l$. 
Note that $0 \leq \aff(\cS_k,\cS_\l)  \leq 1$.  
If $\cS_k$ and $\cS_\l$ intersect in $p$ dimensions, i.e., if $\cS_k \cap \cS_\l$ is $p$-dimensional, then $\cos(\theta_1)=...=\cos(\theta_p)=1$ \cite{golub_matrix_1996}. 
Hence, if $\cS_k$ and $\cS_\l$ intersect in $p\geq 1$ dimensions, we have $\aff(\cS_k,\cS_\l) \geq \sqrt{p/ (\mintwo{d_k}{d_\l}})$. 
We are now ready to state our main result. The corresponding proof is  outlined in Section \ref{sec:proofoutine}. 


\begin{theorem}
Suppose that $\X_\l$ is obtained by choosing $n_\l$ points in $\cS_\l$ at random according to $\vx_j^{(\l)} = \mU^{(\l)} \va^{(\l)}_j, j \in [n_\l]$, where the $\va^{(\l)}_j$ are i.i.d.~uniform on $\US{\d_\l}$, and let $\X = \X_1 \cup ...  \cup  \X_L$. Suppose furthermore that $n_\l/\d_\l \geq 6$ and $d_\l \geq c_2 \log n_\l$, for all $\l\in [L]$, where $c_2$ is a constant that depends on $\d_\l$ only.  If 
\begin{align*}
\max_{k,\l \colon k \neq \l} \aff(\cS_k,\cS_\l) \leq   \frac{1}{15 \log N },
\end{align*}
with $N = |\X|$, then modified TSC yields the correct segmentation of $\X$ with probability at least
$
1 - 3/N - \sum_{\l\in [L]}  \left( n_\l e^{-c(n_\l-1)} + \frac{1}{n_\l^{2}  \log n_\l} \right)
$,
where $c>0$ is a numerical constant.
\label{thm:aff2}
\end{theorem}

Theorem~\ref{thm:aff2} states that modified TSC succeeds with high probability if the affinity between subspaces is sufficiently small, and if the number of points in $\X_\l$ per subspace dimension, i.e., $n_\l/d_\l$, for each $\l$, is sufficiently large. 
Intuitively, we expect that clustering becomes easier when the $n_\l$ increase. To see that Theorem~\ref{thm:aff2}, indeed, confirms this intuition, set $n_\l = n$, for all $\l$, and observe that the probability of success in Theorem~\ref{thm:aff2}, indeed, increases in $n$. 

The original TSC algorithm introduced in \cite{heckel_robust_2013,heckel_subspace_2013} has $q_j = q$, for all points $\vx_j \in \X$, and takes $q$ as an input parameter. 
We note that the statement in Theorem \ref{thm:aff2} applies to this (original) version of TSC as well with the conditions $n_\l/\d_\l \geq 6$ and $d_\l \geq c_2 \log n_\l$ replaced by $\q \leq n_\l/6$ and $\q \geq c_2 \log n_\l$, respectively. 

Theorem \ref{thm:aff2} is proven (for more details see Section \ref{sec:proofoutine}) by showing that the connected components in the graph $G$ with adjacency matrix $\mA$ correspond to the $\X_\l$ with probability satisfying the probability estimate in Theorem \ref{thm:aff2}. 
Previous results for TSC \cite{heckel_subspace_2013} established that each $\vx_i^{(\l)} \in \X_\l$ is connected (in $G$) to other points corresponding to $\X_\l$ only, but it was not shown that the points corresponding to $\X_\l$ form a connected component, which, however, is essential to ensure zero clustering error. 
%

The condition $n_\l/\d_\l \geq 6$ ($\q \leq n_\l/6$ for the original TSC algorithm) is used to establish that each $\vx_j \in \X_\l$ is connected to points corresponding to $\X_\l$ only, while  $d_\l \geq c_2 \log n_\l$ ($\q \geq c_2 \log n_\l$ for the original TSC algorithm) is needed to ensure that subgraphs corresponding to the $\X_\l$ are connected.
%
The latter condition is order-wise necessary. 
%

We finally note that the constant $c_2$ is increasing in $\max_\l \d_\l$. This is likely an artifact of our analysis, as indicated by numerical simulations, not shown here.


\section{Proof outline}
\label{sec:proofoutine}

In the following, we give a brief outline of the proof of Theorem \ref{thm:aff2}. 
For the sake of brevity, we will not detail the minor modifications needed to prove the statement for the original TSC algorithm. 
Let $G$ be the graph with adjacency matrix $\mA$ constructed by the modified TSC algorithm. 
The proof is effected by showing that 
the connected components in $G$ correspond to the $\X_\l$ with probability satisfying the probability estimate in Theorem \ref{thm:aff2}, henceforth simply referred to as ``with high probability''. 
To this end, we first establish that $G$ has no false connections in the sense that the nodes corresponding to $\X_\l$ are connected to nodes corresponding to $\X_\l$ only. 
We then show that, conditional on $G$ having no false connections,  
 the nodes corresponding to $\X_\l$ form a connected subgraph, for all $\l \in [L]$. 

 To establish that $G$ has no false connections, we first show that for each $\vx_j \in \X_\l$ the corresponding set $\S_j(q)$ contains points in $\X_\l$ only, as long as $q \leq n_\l/6$. (The condition $q \leq n_\l/6$ is shown to hold below.) 
This is accomplished through the use of concentration inequalities for order statistics of the inner products between the (random) data points. Specifically,  we show that 
for each $\vx_j^{(\l)} \in \X_\l$, and for each $\X_\l$, we have that
$
z_{(n_\l - \q)}^{(\l)} > \max_{k\neq \l, i} z_{i}^{(k)}
$
with high probability. Here,  $z_{(1)}^{(\l)} \leq z_{(2)}^{(\l)} \leq ...\leq z_{(n_\l-1)}^{(\l)}$ are the order statistics of $\{z_{i}^{(\l)}\}_{i \in [n_\l] \setminus j}$ and $z_{i}^{(k)} = \big| \big< \vx_i^{(k)} ,  \vx_j^{(\l)} \big> \big|$.

We next show that  $q_j$ obtained in Step 1 of the modified TSC algorithm is equal to $d_\l$. 
This is accomplished by establishing that the smallest $q$ for which \eqref{eq:approxadf} holds with $\tau=0$ is $q = d_\l$. Recall that $\mX_{\J_j(q)}$ is the matrix with columns $\vx_i, i \in \S_j(q)$. As long as $q\leq n_\l/6$, $\S_j(q)$ consists of points in $\X_\l$ only (as argued above), therefore $\mX_{\J_j(q)} = \mU^{(\l)} \mA_{\J_j(q)}$, where the columns of $\mA_{\J_j(q)}$ correspond to the $\va_i, i \in \S_j(q)$. Thanks to the orthonormality of $\mU^{(\l)}$, we have
\begin{align}
\norm[2]{ (\mI - \mX_{\J_j(q)} \pinv{\mX}_{\J_j(q)}) \vx_j}
&=\norm[2]{ (\mI - \mA_{\J_j (q)}\pinv{ \mA}_{\J_j(q)}) \va_j }. \label{eq:condadfolooadf}
\end{align}
With probability one, \eqref{eq:condadfolooadf} is strictly positive if $q<d_\l$, and equal to zero if $q=d_\l$, thus $q_j=d_\l$. 
Finally, note that $n_\l/\d_\l \geq 6$ ensures that $q_j\leq n_\l / 6$, which resolves the assumption $q\leq n_\l/6$. 

It remains to show that the nodes corresponding to $\X_\l$ form a connected subgraph, for all $\l \in [L]$. 
Since $\innerprod{\vx_i}{\vx_j} = \innerprod{\va_i}{\va_j}$ for $\vx_i, \vx_j \in \X_\l$, it follows that the subgraph of $G$ corresponding to the points in $\X_\l$ is the $q$-nearest neighbor graph with pseudo-distance metric $\acos(\left|\innerprod{\va_i}{\va_j}\right|)$. 
The proof is then completed using the following result (with $\gamma=3$). 
\begin{lemma}
Let $\va_1,...,\va_n \in \reals^{\d}$ be i.i.d.~uniform on $\US{\d}$, $d>1$, and let $\tilde G$ be the corresponding $\tilde k$-nearest neighbor graph, with $\tilde s(\va_i, \va_j) = \acos( \left|\innerprod{\va_i}{\va_j}\right| )$ as the underlying distance metric. Then, with $\tilde k = \gamma  \,c_1\log n$, where $c_1$ depends on $d$ only, and is increasing in $\d$, for every $\gamma > 0$,  we have
$
\PR{\tilde G \text{ is connected}  } \geq 1  -   \frac{2}{n^{\gamma-1}   \gamma \log n}.
$
\label{lem:connectivityknng}
\end{lemma}
%


\section{Numerical results}

We compare modified TSC to TSC, SSC, and SSC-OMP  on synthetic and on real data. For SSC, we use the implementation in \cite{elhamifar_sparse_2013}.

\begin{figure}
\begin{center}
\begin{tikzpicture}[scale=0.8]
    \begin{axis}[
        	xlabel=$n$,
	ylabel=clustering error,
	]





%

        \addplot[color=black,mark=none] coordinates {
(5, 5.200000e-01)
(10, 4.775000e-01)
(15, 3.916667e-01)
(20, 3.750000e-01)
(25, 2.140000e-01)
(30, 1.516667e-01)
(35, 9.571429e-02)
(40, 6.187500e-02)
(45, 5.055556e-02)
(50, 3.200000e-02)
(55, 2.181818e-02)
(60, 1.875000e-02)
(65, 1.653846e-02)
(70, 1.964286e-02)
(75, 1.466667e-02)
(80, 1.187500e-02)
(85, 7.941176e-03)
(90, 7.500000e-03)
(95, 8.421053e-03)
(100, 6.000000e-03)
(105, 5.238095e-03)
}; 

\addlegendentry{TSC};

              \addplot[color=black,dashdotted,mark=none]    
                 coordinates {
(5, 5.050000e-01)
(10, 4.950000e-01)
(15, 4.066667e-01)
(20, 2.962500e-01)
(25, 1.550000e-01)
(30, 1.008333e-01)
(35, 7.142857e-02)
(40, 2.687500e-02)
(45, 3.666667e-02)
(50, 2.050000e-02)
(55, 1.545455e-02)
(60, 1.500000e-02)
(65, 1.115385e-02)
(70, 1.071429e-02)
(75, 1.166667e-02)
(80, 7.187500e-03)
(85, 5.882353e-03)
(90, 4.444444e-03)
(95, 6.315789e-03)
(100, 3.250000e-03)
(105, 2.142857e-03)
}; 
\addlegendentry{modified TSC};

                \addplot[color=black,loosely dashed]    
                 coordinates {
(5, 5.400000e-01)
(10, 4.450000e-01)
(15, 2.233333e-01)
(20, 6.000000e-02)
(25, 3.400000e-02)
(30, 1.750000e-02)
(35, 3.571429e-03)
(40, 3.125000e-03)
(45, 3.888889e-03)
(50, 5.000000e-04)
(55, 1.363636e-03)
(60, 4.166667e-04)
(65, 1.153846e-03)
(70, 0)
(75, 0)
(80, 6.250000e-04)
(85, 2.941176e-04)
(90, 2.777778e-04)
(95, 2.631579e-04)
(100, 2.500000e-04)
(105, 2.380952e-04)
}; 
\addlegendentry{SSC-OMP};

                \addplot[color=black,dotted]    
                 coordinates {
(5, 5.100000e-01)
(10, 4.025000e-01)
(15, 1.483333e-01)
(20, 2.125000e-02)
(25, 4.000000e-03)
(30, 2.500000e-03)
(35, 7.142857e-04)
(40, 0)
(45, 5.555556e-04)
(50, 0)
(55, 0)
(60, 0)
(65, 3.846154e-04)
(70, 0)
(75, 0)
(80, 0)
(85, 0)
(90, 0)
(95, 0)
(100, 0)
(105, 0)
}; 
\addlegendentry{SSC};

\end{axis}
  
\end{tikzpicture}  
\end{center}
\vspace{-0.6cm}
\caption{\label{fig:syntcomp} Clustering error as a function of the number of points $n$ in each subspace.}
\end{figure}
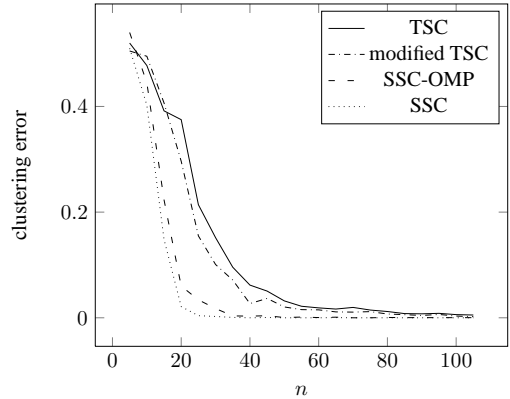

\vspace{-0.3cm}
\paragraph*{Synthetic data:}
We generate $L=8$ subspaces of $\reals^{120}$ with dimension $d=30$ each. 
Specifically, we choose the corresponding $\mU^{(\l)} \in \reals^{m\times d}$ uniformly at random from the set of all orthonormal matrices in $\reals^{m\times d}$, with the first $d/3 = 10$ columns being equal. This ensures that the subspaces intersect in at least $d/3$ dimensions and hence $\aff(\cS_k,\cS_\l) \geq 1/\sqrt{3}$. 
The points corresponding to $\cS_\l$ are chosen at random according to $\vx_j^{(\l)} = \mU^{(\l)} \va^{(\l)}_j  + \ve^{(\l)}_j  , j \in [n]$, where the $\va^{(\l)}_j$ are i.i.d.~uniform on $\US{\d}$ and the $\ve^{(\l)}_j$ are i.i.d. $\mathcal N( \mathbf 0, (\sigma^2/m)  \mI_m)$ with $\sigma^2 = 0.3$. For each $n$, the clustering error is averaged over $50$ problem instances. 
We choose $q=20$ for TSC, stop OMP in OMP-SSC after $20$ iterations, and set $\tau = 0.45$ in modified TSC. 
The results, summarized in Fig.~\ref{fig:syntcomp}, show that SSC and SSC-OMP outperform TSC and modified TSC. However, TSC is computationally less demanding. Finally, modified TSC is seen to perform slightly better than the original TSC algorithm.


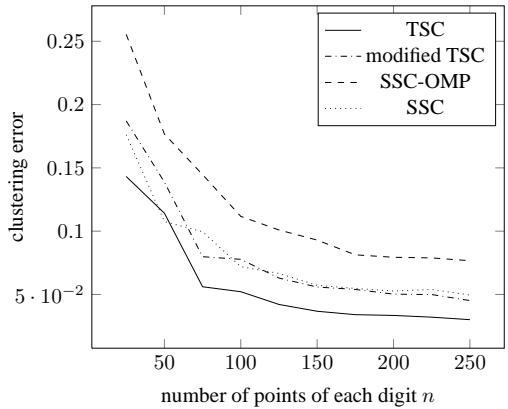
\begin{figure}
\begin{center}
\begin{tikzpicture}[scale=0.8]
    \begin{axis}[
        	xlabel=number of points of each digit $n$,
	ylabel=clustering error,
	]





%
        \addplot[color=black,error bars/.cd,
    y dir=both,y explicit,] coordinates {
(25, 1.432000e-01)
(50, 1.142000e-01)
(75, 5.608889e-02)
(100, 5.216667e-02)
(125, 4.210667e-02)
(150, 3.673333e-02)
(175, 3.403810e-02)
(200, 3.343333e-02)
(225, 3.205926e-02)
(250, 3.009333e-02)
}; 
\addlegendentry{TSC};
        \addplot[color=black, dashdotted] coordinates {
(25, 1.870667e-01)
(50, 1.388000e-01)
(75, 7.982222e-02)
(100, 7.776667e-02)
(125, 6.296000e-02)
(150, 5.586667e-02)
(175, 5.396190e-02)
(200, 5.025000e-02)
(225, 4.992593e-02)
(250, 4.517333e-02)
}; 
\addlegendentry{modified TSC};

                \addplot[color=black,dashed,error bars/.cd,
    y dir=both,y explicit,
                ]    
                 coordinates {
(25, 2.554667e-01)
(50, 1.767333e-01)
(75, 1.446667e-01)
(100, 1.117000e-01)
(125, 1.009333e-01)
(150, 9.295556e-02)
(175, 8.129524e-02)
(200, 7.936667e-02)
(225, 7.887407e-02)
(250, 7.665333e-02)
}; 
\addlegendentry{SSC-OMP};

                \addplot[color=black,dotted,error bars/.cd,
    y dir=both,y explicit,
                ]    
                 coordinates {
(25, 1.761333e-01)
(50, 1.076000e-01)
(75, 9.955556e-02)
(100, 7.196667e-02)
(125, 6.661333e-02)
(150, 5.722222e-02)
(175, 5.476190e-02)
(200, 5.270000e-02)
(225, 5.395556e-02)
(250, 4.949333e-02)
}; 
\addlegendentry{SSC};

\end{axis}
  
\end{tikzpicture} 
\vspace{-0.6cm} 
\end{center}

\caption{\label{fig:compssctsc}  Empirical mean and standard deviation of the clustering error for clustering handwritten digits.}
\vspace{-0.2cm} 
\end{figure}

\vspace{-0.3cm}
\paragraph*{Clustering handwritten digits:}
We next consider the problem of clustering handwritten digits. Specifically, we work with the MNIST data set of handwritten digits \cite{mnist_2013},  and use the test set that contains 10,000 centered $28\times 28$ pixel images of handwritten digits. The assumption underlying the idea of posing this problem as a subspace clustering problem is that the vectorized images of the different handwritten versions of a single digit lie in a low-dimensional subspace of unknown dimension and orientation. 
The empirical mean and variance of the corresponding clustering errors, depicted in Fig.~\ref{fig:compssctsc}, are computed by averaging over 100 of the following problem instances. 
We choose the digits $\{0,2,4,8\}$ and for each digit we choose $n$ vectorized and normalized images uniformly at random from the set of all images of that digit. We choose $q=7$ for TSC,  stop OMP in OMP-SSC after $7$ iterations, and use $\tau = 0.45$ in modified TSC. 
The results show that TSC outperforms modified TSC, SSC, and SSC-OMP. 
TSC outperforming modified TSC may be attributed to the fact that for this dataset $q_j$ is large for several $j$, which means that some digits can not be well represented by its nearest neighbors. 
We hasten to add that for other problems and datasets, SSC may outperform TSC as, e.g., for the problem of clustering faces. 

\paragraph*{Acknowledgments:} We would like to thank Mahdi Soltanolkotabi for helpful and inspiring discussions.



%
%
%

\end{document}